\documentclass[runningheads]{llncs}
\usepackage{xfrac}    

\usepackage{color}
\usepackage{multicol}
\usepackage{footmisc}
\usepackage{epstopdf}
\usepackage{subcaption}
\def\acknowledgement{\par\addvspace{17pt}\small\rmfamily
\trivlist\if!\ackname!\item[]\else
\item[\hskip\labelsep 
{\bfseries\ackname}]\fi}

\newenvironment{acknowledgements}{\begin{acknowledgement}}
{\end{acknowledgement}}

\begin{document}
\title{Simulating Crowds and Autonomous Vehicles}
%
%
\author{John Charlton\inst{1}\orcidID{0000-0001-8402-6723} \and
Luis Rene Montana Gonzalez\inst{1} \orcidID{0000-0002-0511-0466} \and
Steve Maddock\inst{1}\orcidID{0000-0003-3179-0263} \and
Paul Richmond\inst{1}\orcidID{0000-0002-4657-5518}}
\authorrunning{J. Charlton et al.}
%
\institute{University of Sheffield, Sheffield S10 2TN, UK \email{\{j.a.charlton,lrmontanagonzalez1,s.maddock,p.richmond\}@sheffield.ac.uk} }
\maketitle              
\begin{abstract}

Understanding how people view and interact with autonomous vehicles is important to guide future directions of research. One such way of aiding understanding is through simulations of virtual environments involving people and autonomous vehicles. We present a simulation model that incorporates people and autonomous vehicles in a shared urban space. The model is able to simulate many thousands of people and vehicles in real-time. This is achieved by use of GPU hardware, and through a novel linear program solver optimized for large numbers of problems on the GPU. The model is up to 30 times faster than the equivalent multi-core CPU model. 
\end{abstract}
%

\section{Introduction}

Self-driving vehicles are an important area of current research and industry. There is a lot of work examining how best to implement this developing technology. One guiding factor of this is an understanding of how people feel about the use of autonomous vehicles, since they will be users, as passengers, and will also interact with the vehicles as pedestrians. One way of gaining such understanding is through the use of visualizations of virtual simulations of pedestrian crowds and autonomous vehicles. Creating simulations that combine people and autonomous vehicles demonstrates potential phenomena that may arise from this novel situation. It also provides outreach opportunities to gauge public opinion, and increase confidence of sharing the same area. Of special interest is the use of autonomous vehicles that share the same space as pedestrians. This type of transportation increases the area available for both the vehicles and people to move around in. It also allows the autonomous vehicles to reach any location, resulting in true door-to-door transportation. This could potentially allow many more people and vehicles to be able to use the same space compared to segregating them, resulting in larger overall accessible areas, potentially reducing traffic and densities.

Part of the requirement for such simulations is the ability to simulate many people and vehicles. This allows for large public spaces containing many thousands of people to be visualized, such as areas surrounding busy transport hubs or city centers. Another requirement is to run in real-time. This allows for user interactivity, particularly for applications such as virtual reality. The choice of simulation model tends to be one in which the more computationally complex it is, the more realistic the motion it produces tends to be. By using particular agent-based simulation models with inherent parallelism, it is possible to create simulations with greater numbers of people and agents, running at faster speeds. This is achieved by running such simulations on GPU hardware, due to the larger throughput available for GPUs compared to CPUs. This has been seen in other pedestrian models previously \cite{barut_combining_2018,bleiweiss_multi_2009}.

The work presented here builds on previous work by Charlton et al. \cite{charlton_fast_2019}, which presented an efficient pedestrian steering model for GPU hardware using the ORCA steering model. That work shows speed increases of up to 30 times compared to the original multi-core CPU model. The speedup is achieved by parallelizing as much of the data and computation as possible, choosing data parallel algorithms and spatial partitioning to allow communication between people to provide speedup. It makes use of a  low-dimensional linear program solver that efficiently solves numerous problems simultaneously on the GPU \cite{charlton_two-dimensional_2019}, and adapts the solver to optimize for a position close to a point in 2D space, rather than a linear function as described in the paper. This model also uses a grid-based spatial partitioning scheme to efficiently communicate data within the GPU \cite{richmond_flame_2011}.

The work presented in this paper expands upon the work of Charlton et al. \cite{charlton_fast_2019} by including autonomous vehicles in the simulation. The model has been extended with rules to explain interaction between people and vehicle and between vehicle and vehicle. It allows high performance simulations involving both people and autonomous vehicles interacting in the same space. The resulting model has been incorporated into a visualization to demonstrate the effects of simulations involving people and autonomous vehicles in the same shared space.

The organization of the paper is as follows. Section \ref{sec:background} covers background information and related work. Section \ref{sec:algorithmology} explains in detail the implementation of the ORCA model on the GPU and extensions to allow for autonomous vehicles. Section \ref{sec:results} presents results of the visualization of the autonomous vehicles and people simulation, and discussion of the multi-core CPU and GPU ORCA models. Finally, section \ref{sec:conclusion} gives conclusions.

\section{Background} \label{sec:background}

Many types of models have been proposed to generate local pedestrian motion and collision avoidance \cite{pettre_motion_2008,bousseau_introduction_2017,thalmann_populating_2006}. The simplest separation of steering models is between continuum models and microscopic models. Continuum models attempt to treat the whole crowd in a similar way to a fluid, allowing for fast simulation of larger numbers of people, but are lacking in accuracy at the individual person scale \cite{narain_aggregate_2009}.  Moving part of the calculation to the GPU has shown performance improvements \cite{fickett_GPU_2007}. Overall, however, the model is not ideal for solving on the GPU due to the large sparse data structures. In comparison, microscopic models tend to be paired with a global path planner to give people goal locations and trajectories. Such models specify rules at the individual person scale, with crowd-scale dynamics being an emergent effect of the rules and interactions, and easily allow for non-homogeneous agents and behavior. 

Popular microscopic models are cellular automata (CA), social forces \cite{helbing_social_1995} and velocity obstacles (VO) \cite{fiorini_motion_1998}. CA are popular due to the ability to reproduce observable phenomena \cite{blue_emergent_1998,blue_cellular_1999}, but a downside is the inability to reproduce other behaviors due to using discrete space. CA models are computationally lightweight and lend themselves well to specify certain complex behavior. However, CA pedestrian models tend to use discrete spatial rules, where the order of agent movements are sequential, which does not lend itself to parallelism and GPU implementations \cite{schonfisch_synchronous_1999}. Social force models use a computationally lightweight set of rules that allows for crowd-scale observables such as lane formation. They are well suited to parallelizing on the GPU since all agents can be updated simultaneously, with good performance for many simulated people \cite{karmakharm_agent-based_2010,richmond_high_2008}. However, generated simulations can result in unrealistic looking motion and produce undesirable behavior at large densities.

Velocity obstacles (VO) work by examining the velocity and position of nearby moving objects to compute a collision-free trajectory. Velocity-space is analyzed to determine what velocities can be taken which do not cause collisions. 
VO models lend themselves to parallelization since agents are updated simultaneously and navigate independently of one another with minimal explicit communication. It tends to be more computational and memory intensive than social forces models, but the large throughput capability of the GPU for such parallel tasks make it a very suitable technique for GPU implementation.
Early models assumed that each person would take full responsibility for avoiding other people. Several variations include the reactive behavior of other models \cite{abe_collision_2001,kluge_recursive_2006,fulgenzi_dynamic_2007}. One example is reciprocal velocity obstacles (RVO), where the assumption is that all other people will take half the responsibility for avoiding collisions \cite{berg_reciprocal_2008,guy_clearpath:_2009}. This model has been implemented on the GPU \cite{bleiweiss_multi_2009} and has shown credible speedup over the multi-core CPU implementation through use of hashing instead of naive nearest neighbor search. Group behavior has also been included in VO models \cite{he_dynamic_2016,yang_proxemic_2016} allowing people to be joined into groups. Such people attempt to remain close to other members of the group and aim for the same goal location.
25There are many further extentions to VOs to specialize movement and behaviors to account for a variety of agent and robot types, such as elliptical agents \cite{best_real-time_2016}, holonomic movement \cite{hughes_holonomic_2014}, and non-holonomic movement \cite{wang2018reciprocal}\cite{huang2019generalized}. 
A further extension of particular interest is optimal reciprocal collision avoidance (ORCA). It provides sufficient conditions for collision-free motion. It works by solving low-dimension linear programs. Freely available code libraries have been implemented for both single- and multi-core CPU \cite{snape_optimal_2019}.

VO techniques are very suitable candidates for GPU implementation. The RVO model and implementation by Bleiweiss \cite{bleiweiss_multi_2009} show notable performance gains against multi-core CPU equivalent models. However, these methods must perform expensive calculations to find a suitable velocity. They tend to perform slower and are not guaranteed to find the best velocity. ORCA is deemed more suitable because of its performance relative to other VO models and collision-free motion, theoretically providing ``better" motion (i.e. less collisions). A critical analysis of common VO approaches is presented by Douthwaite et al. \cite{douthwaite2019velocity}. In their analysis they find the ORCA model to scale the best out of the tested VO models, and also find ORCA provides the smoothest trajectories.  

Linear programming is a way of maximizing an objective function subject to a set of constraints. For ORCA, linear programming is used to find the closest velocity to a person's desired velocity which does not result in collisions. It is important to choose a solver that is efficient on the GPU at low dimensions. A popular solver type is the Simplex method. This is best suited for large dimension problems and struggles at lower dimensions. The incremental solver \cite{seidel_small-dimensional_1991} is efficient at low dimensions but suffers on the GPU due to load balance: not all GPU threads have the same amount of computation, which reduces the performance on such parallel architecture.
The batch GPU two-dimension linear solver \cite{charlton_two-dimensional_2019} is an efficient way to solve the numerous linear problems simultaneously. We make use of this approach, demonstrating its use for large-scale simulations.

There is a lot of work examining the ethics and safety of autonomous vehicles \cite{gogoll_autonomous_2017,lin_why_2016,mcbride_ethics_2016,althoff_comparison_2011,guo_is_2019}. There is also examination into understanding how users respond to autonomous vehicles, and therefore how such vehicles should be programmed. A recent survey and overview of this is provided by Rasouli and Tsotsos \cite{rasouli_autonomous_2019}. Schwarting et al. use a variety of psycological metrics which model the altruism and selfishness of people to predict their interactions towards autonomous vehicles \cite{schwarting_social_2019}. Bonnefon et al. examined the idea of pedestrian-first safely compared to passenger-first safety of autonomous vehicles. That is, given a scenario which will cause harm to either the pedestrians or passengers, whose safety will be prioritized. They found people would prefer others to buy pedestrian-first safe vehicles, but would prefer to ride in passenger-first safe vehicles, which provides a moral, ethical, and utilitarian dilemma as to the choice of algorithms used in autonomous vehicles \cite{bonnefon_social_2016}. Pettersson and Karlsson \cite{pettersson_setting_2015} find that different ways of presenting information on the same subject can yield different types of data. This means that the larger variety of methods of presenting and interacting with information about autonomous vehicles, the more understanding of the users can be found. 
Work into simulating autonomous vehicles has been carried out with agent-based models. Boesch and Ciari \cite{boesch_agent-based_2015} present an agent-based simulation model for autonomous vehicles, called MATsim, with the aim of it being simple enough for usage for those interested. This tool is used in a variety of applications, such as examining an autonomous taxi service \cite{horl_agent-based_2017}, providing a theoretical examination in which to target future experimental research.
Zhang et al. \cite{zhang_exploring_2015} try to predict the effect autonomous vehicle will have on city parking through use of an agent based model, and estimate that up to 90\% of parking demand could be eliminated.

\section{The Algorithm} \label{sec:algorithmology}

This section provides an overview of the algorithm as well as important changes to allow simulating of autonomous vehicles. For more in-depth description of the original ORCA algorithm, see the work of van den Berg et al. \cite{berg_reciprocal_2011}, or for more detail on the GPU implementation, see the work of Charlton et al. \cite{charlton_fast_2019}. The main changes are changes to the avoidance rules determining how much people and vehicles should avoid each other.

\begin{figure}
    \centering
    \begin{subfigure}{0.45\textwidth}
        \includegraphics[width=0.9\linewidth]{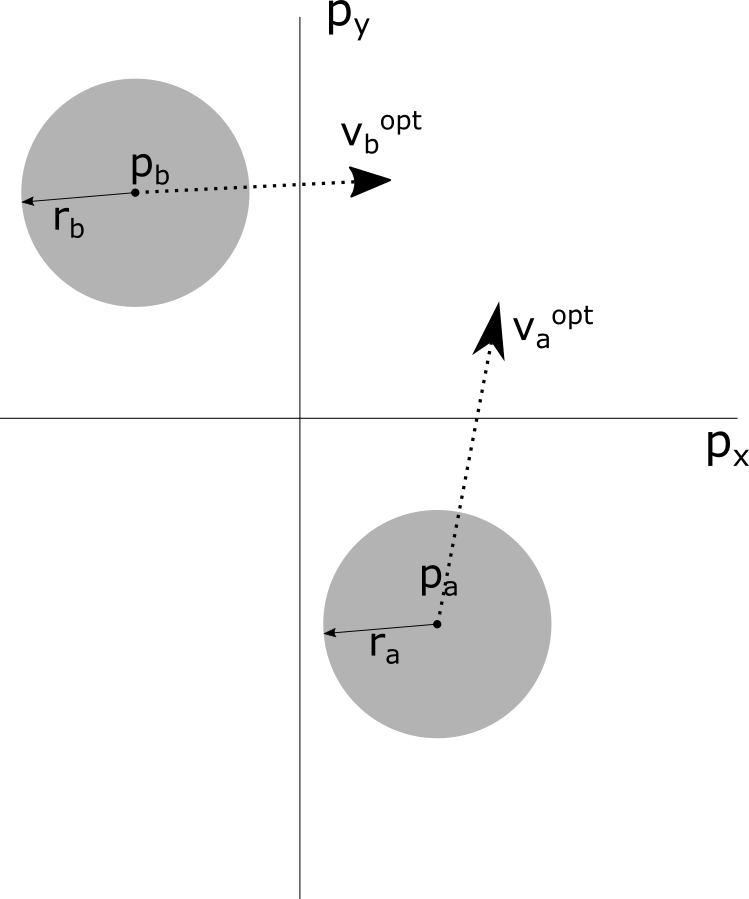}
        \caption{ }
    \end{subfigure}
     \begin{subfigure}{0.45\textwidth}
        \includegraphics[height=6cm]{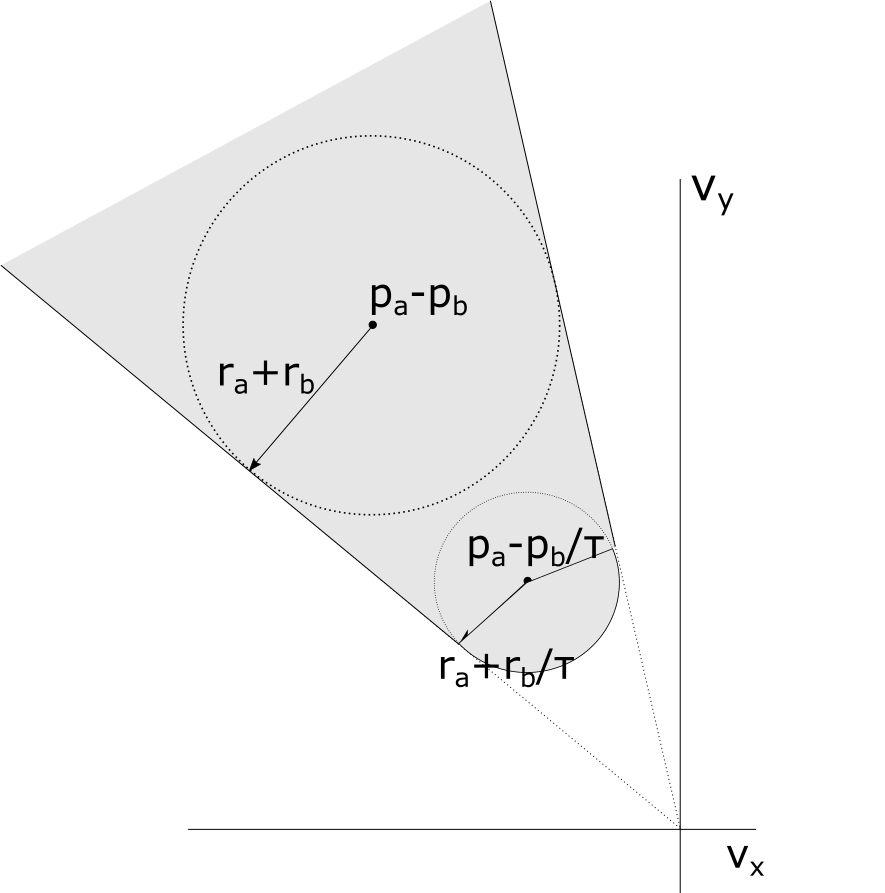}
        \caption{ }
    \end{subfigure}
     \begin{subfigure}{0.45\textwidth}
        \includegraphics[width=0.9\linewidth]{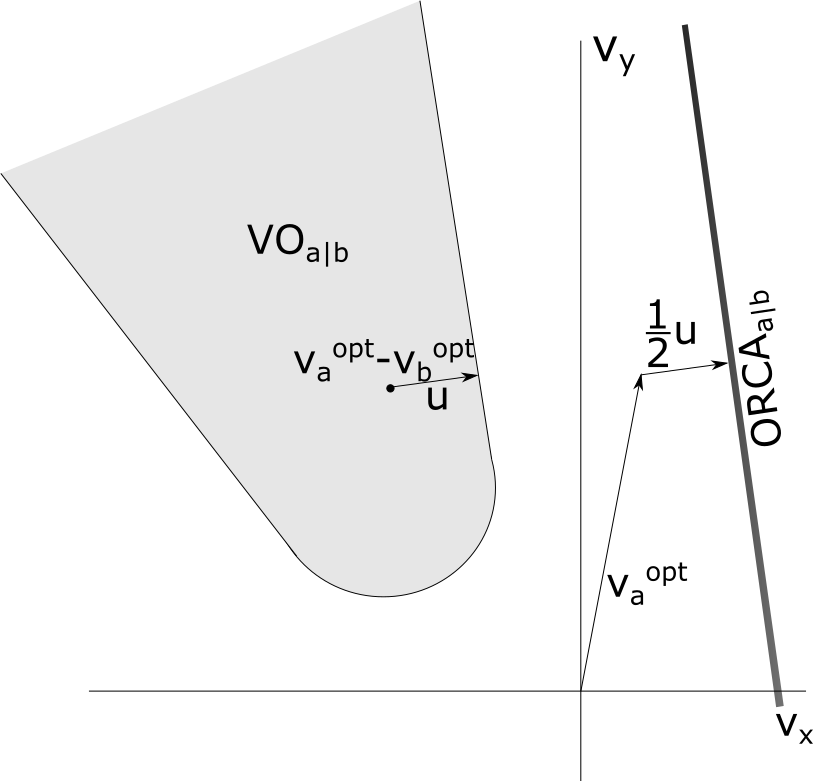}
        \caption{}
        \label{fig:ORCAhalfu}
    \end{subfigure}
    \begin{subfigure}{0.45\textwidth}
        \includegraphics[width=0.9\linewidth]{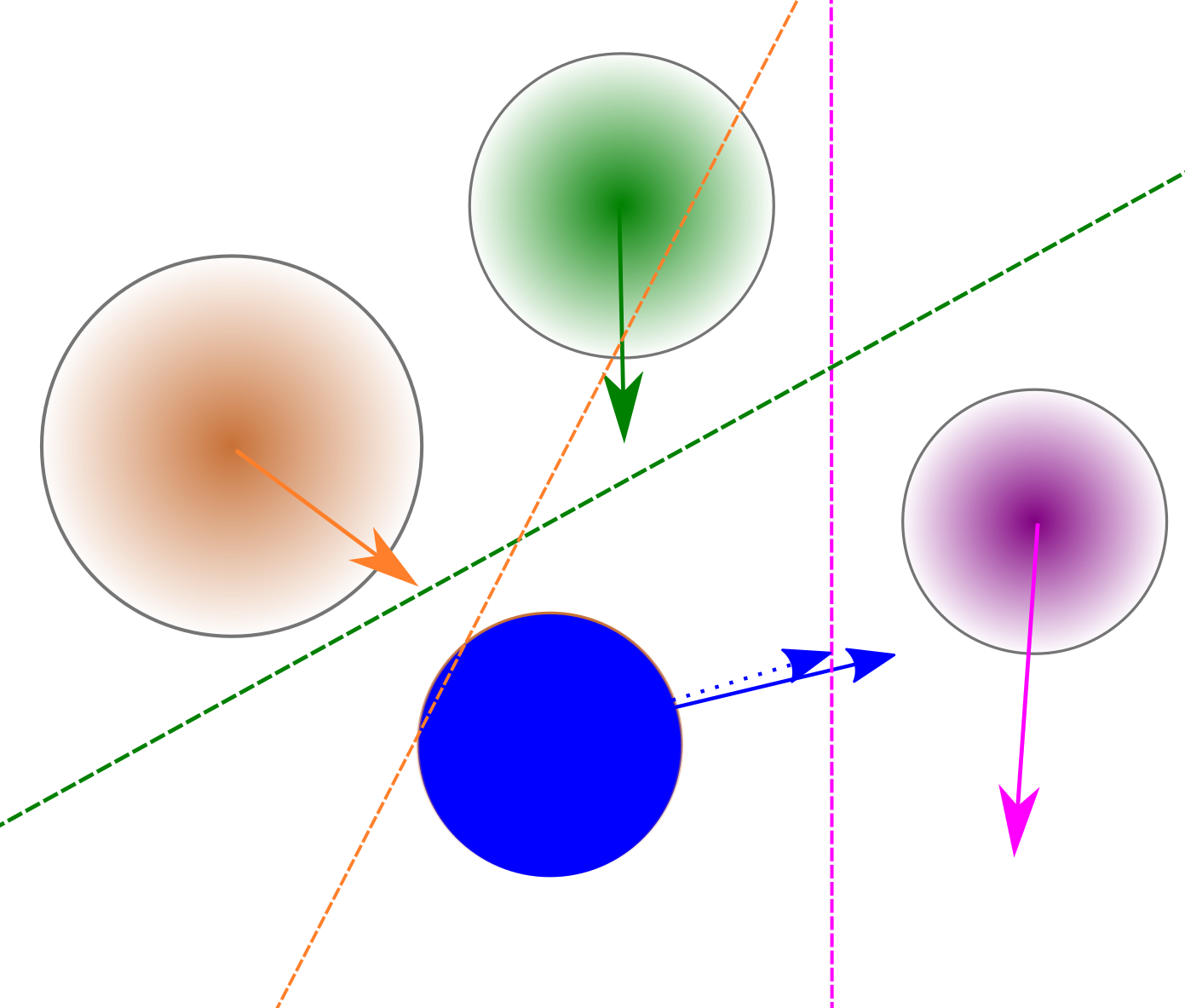}
        \caption{ }
        \label{fig:ORCAsolve}
    \end{subfigure}
    \caption{(a) A system of 2 people $a$ and $b$ with corresponding radius $r_a$ and $r_b$. (b) The associated velocity obstacle $VO_{a|b}$ in velocity space for a look-ahead period of time $\tau$ caused by the neighbor $b$ for $a$. (c) The vector of velocities $v_a^{opt} - v_a^{opt}$ lies within the velocity obstacle $VO_{a|b}$. The vector $u$ is the shortest vector to the edge of the obstacle from the vector of velocities. The corresponding half-plane $ORCA_{a|b}$ is in the direction of $u$, and intersects the point $v_a^{opt} + $ \sfrac{1}{2} $u$. (d) A view of a blue agent and its neighbors, as well as the generated half-planes caused by the neighbors interacting with the blue agent. The solid blue arrow shows the desired velocity of the blue agent. The dotted blue arrow is the resulting calculated velocity that does not collide with any neighbor in time $\tau$.}
    \label{fig:ORCAlines}
\end{figure}

As an overview to the ORCA model, each agent in the model has a start location and an end location they want to reach as quickly as possible, subject to an average speed and capped maximum speed. For each simulation iteration, each agent ``observes" properties of other nearby agents, namely radius, the current position and velocity. For each nearby agent a half-plane of restricted velocities is calculated (figure \ref{fig:ORCAlines}). By selecting a velocity not restricted by this half-plane, the two agents are guaranteed to not collide within time $\tau$, where $\tau$ is the \textit{lookahead time}, the amount of forward time planning people make to avoid collisions. By considering all nearby agents, the set of half-planes creates a set of velocities that, if taken, do not collide with any nearby agents in time $\tau$. The agent then selects from the permissible velocities the one closest to its desired velocity and goal. Figure \ref{fig:ORCAsolve} shows the resulting half-planes caused by neighboring agents on an example setup, and the optimal velocity that most closely matches the person's desired velocity.

It is possible that the generated set of half-planes does not contain any possible velocities. Such situations are caused by large densities of people. The solution is to select a velocity that least penetrates the set of half-planes induced by the other agents. In this case, there is no guarantee of collision-free motion. 

The computation of velocity subject to the set of half-planes is done using linear programming. The problem for the linear program is defined with the constraints corresponding to the half-plane $ORCA_{a|b}$ of velocities, attempting to minimize the difference of the suitable velocity from the desired velocity. Since each agent needs to find a new velocity, there is a linear problem corresponding to each agent, each iteration. The algorithm used to solve this is the batch-GPU-LP algorithm \cite{charlton_two-dimensional_2019}. It is an algorithm designed for solving multiple low-dimensional linear programs on the GPU, based on the randomized incremental linear program solver of Seidel \cite{seidel_small-dimensional_1991}. 

This batch-LP solver works by initially assigning each thread to a problem (i.e. one pedestrian). Each thread must solve a set of half-plane constraints, subject to an optimization function. Respectively, these are that the person should not choose a velocity that collides with other people, and the person wants to travel as close to their desired velocity as possible.

Each half-plane constraint is considered incrementally. If the current velocity is not satisfied by the currently considered constraint a new valid velocity is calculated. The calculation of a new velocity is one of the most computationally expensive operations. It is also very branched, as only only some of the solvers require a new valid velocity and others can maintain their current value. This branching calculation causes the threads that do no need to perform a calculation to remain idle while the other threads perform the operation. This is an unbalanced workload on the GPU device and can vastly reduce the throughput as many threads do not perform any calculations, exacerbated by the fact that those threads performing the operation must take a lot of time to complete the operation.

The implementation of this calculation uses ideas from cooperative thread arrays \cite{wang_gunrock:_2016} to subdivide the calculation into \textit{``work units"}, blocks of equal size computation. These work units can be transferred to and computed by different threads, allowing for a balanced work load and good performance. If the thread does not need to compute a new velocity, then it can aid in another problem's calculation. This algorithm shows performance improvements over state-of-the-art CPU LP solvers and other GPU LP solvers \cite{charlton_two-dimensional_2019}.

\subsection{Extension}

This section presents the changes made to the ORCA model to allow for simulating both people and autonomous vehicles. ORCA is a version of an RVO model. In this model, people will take 50\% of the responsibility to avoid colliding. This is reflected in the calculation of the half plan $ORCA_{A|B}$ being shifted by \sfrac{1}{2} $u$, explained in Figure \ref{fig:ORCAhalfu}. Because the agent takes (at least) half of the responsibility for avoidance, it will assume the other agent will take the remaining half, and hence collision will be avoided between the two agents.

We include a variable parameter $f_{A|B}$, that represents the fractional responsibility that agent A will take to avoid agent B. This variable is used when calculating the position of the half-plane $ORCA_{A|B} = v_{opt} + f_{A|B} u$. For the standard case of people-people interaction, $f=$\sfrac{1}{2}.

Because of the different nature of movement of people and vehicles, the avoidance taken between vehicles and people will not be reciprocal. As such, the weighting of $u$ in Figure \ref{fig:ORCAlines} will be different. Since colliding with an autonomous vehicle could provide greater risk of health and damage, we assume people will tend to naturally take more responsibility in avoiding collisions with vehicles (the amount of avoidance is $f_{person|car} > $  \sfrac{1}{2}). 

The vehicles themselves could move in a variety of ways. In order to maintain the parallelism of the ORCA model, we let autonomous vehicles follow the same sets of steering rules as people, i.e. they also follow the ORCA algorithm. By ensuring the executing code path is equivalent for all agents, regardless of size or how much they should contribute to avoiding others, there is no performance loss by varying the number of vehicles in the simulation, compared to simulating only people. When a vehicle must avoid colliding with another car, they will reciprocally avoid each other and each take half the responsibility, much as people do when interacting with people, i.e. $f_{car|car}=$\sfrac{1}{2}. 

To ensure guaranteed collision-free motion, $f_{A|B} + f_{B|A} >= 1$. If an agent knows how much another will avoid collisions, they can know how much they should avoid collision. If a person knows that vehicles will not attempt to avoid collisions (i.e. $f_{car|person} = 0$), then people know they are entirely responsible for getting out of the way ($f_{person|car}=1$). 

Regarding the code itself, it has been rewritten to minimize branching code paths which ensures GPU performance is maintained, regardless of the proportion of people and autonomous vehicles in the simulation. The GPU is composed of many threads, much like multicore CPUs, though with many more threads. These threads are grouped into execution units called warps. A warp usually consists of 32 threads. If threads in a warp execute different code the other threads, some are masked out and remain idle while the branching code is executed, and hence reduce the performance of the code. The code is written to ensure both people and autonomous vehicles follow the same code path, and so no threads are idle for this reason, and performance is not affected by the inclusion, or proportion of autonomous vehicles.

\section{Results} \label{sec:results}
This section presents the results of two experiments. The first experiment is composed of two test cases to demonstrate the appearance and correctness of the model. The first test case is a 2-way crossing and the second test case is an 4-way crossing. The second experiment demonstrates the performance compared to the equivalent multi-core CPU version \cite{berg_reciprocal_2011,snape_optimal_2019}.

For the first experiment, all the test cases are set up in a similar way. Multiple associated start and end regions are chosen, such that people and vehicles are spawned in a start region with a target in an associated end region. Random spawn locations are chosen so that there is no overlap with other people within a certain time period based on person size and speed. Within a simulation, each agent has a goal location to aim for. The agent's velocity is in the direction of the goal location, scaled to the walking speed. Once an agent reaches the goal location they are removed from the simulation. Once all agents have reached their goal the simulation is ended. The avoidance responsibility values are set such that people are entirely responsible for avoiding collisions with the autonomous vehicles. This is mathematically represented as $f_{person|car} = 1$, $f_{car|person} = 0$.

\begin{figure}
\centering
    \begin{subfigure}[t]{1\textwidth}
      \includegraphics[width=1\textwidth]{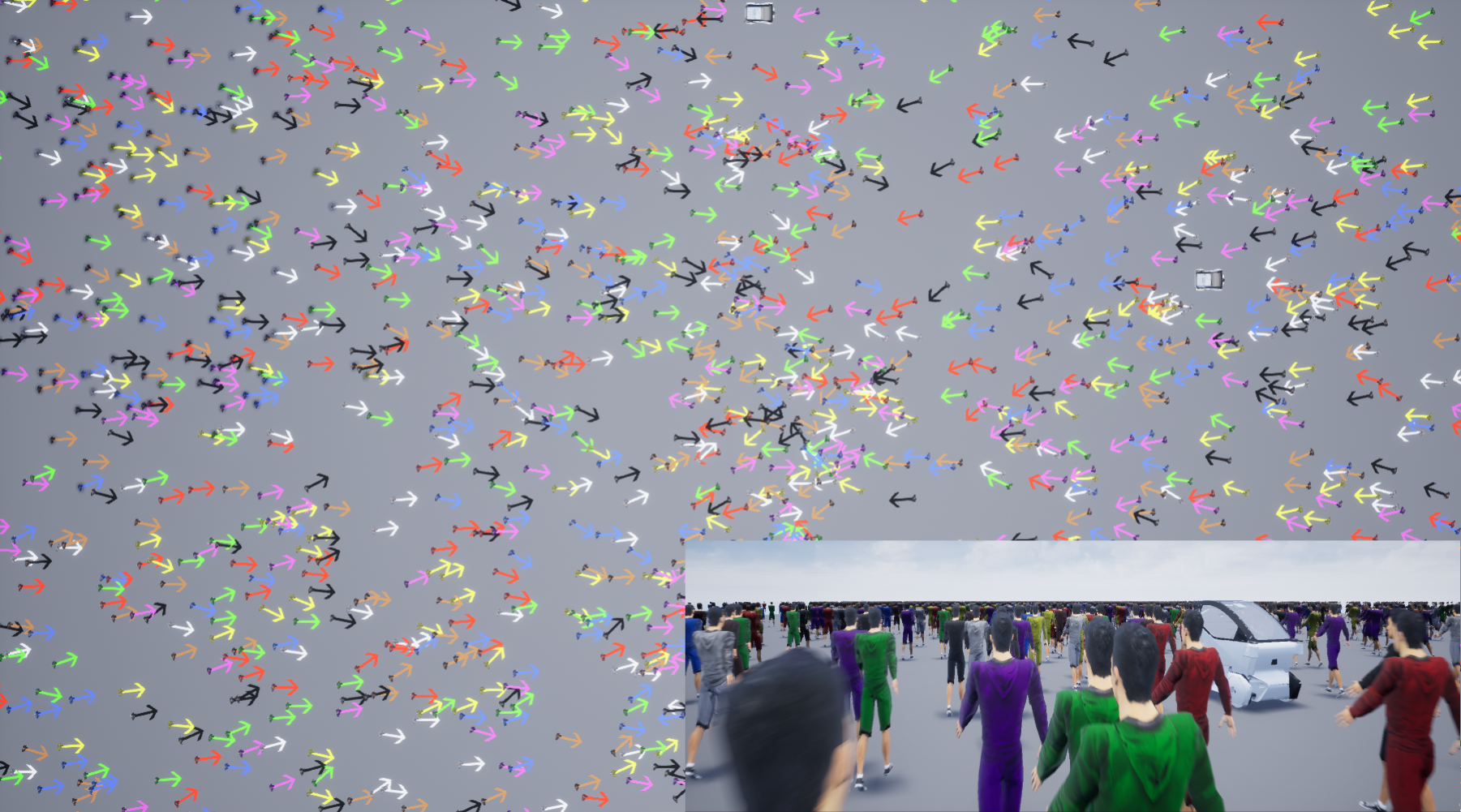}
    \end{subfigure}%
    \caption{Visualization of 2,500 agents in Unreal. Two crowds navigate past each other, one heading from left to right and the other heading from right to left. Inset: an eye-level view of the simulation, taken at the same time}
    \label{fig:vis1}
\end{figure}

\begin{figure}
    \centering
    \begin{subfigure}{1\textwidth}
        \includegraphics[width=0.9\linewidth]{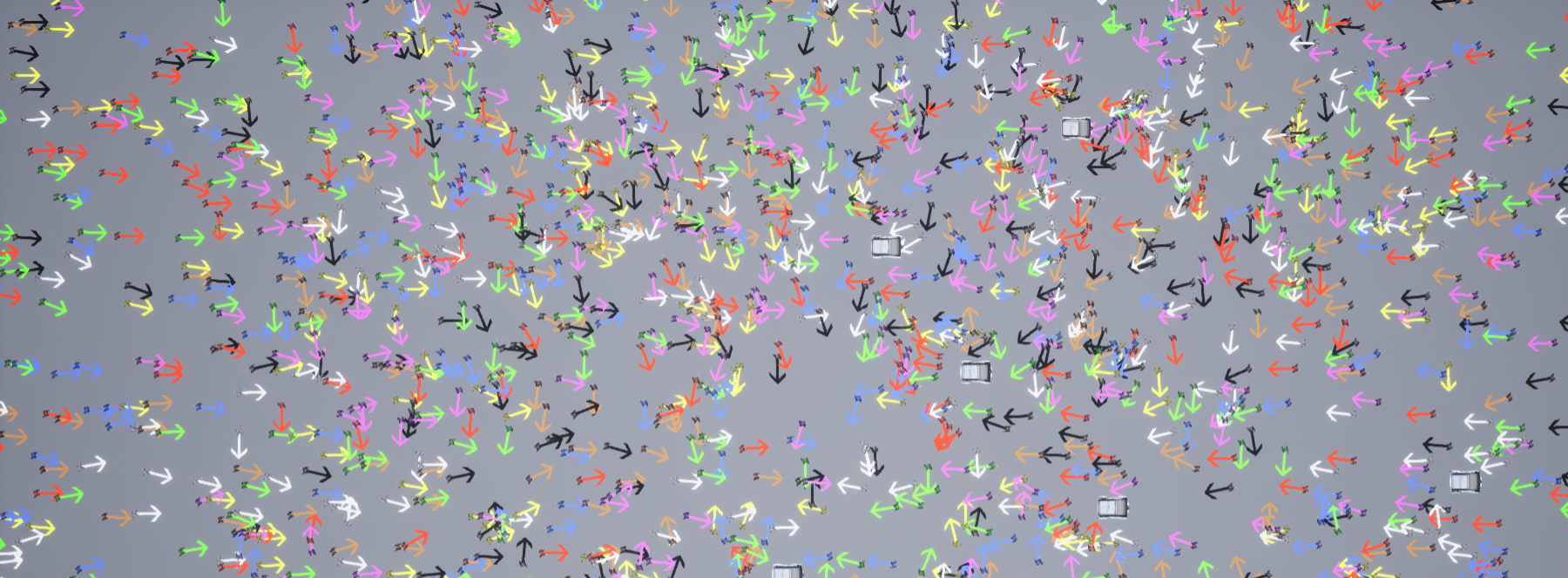}
        \caption{ }
    \end{subfigure}
     \begin{subfigure}{1\textwidth}
        \includegraphics[width=0.9\linewidth]{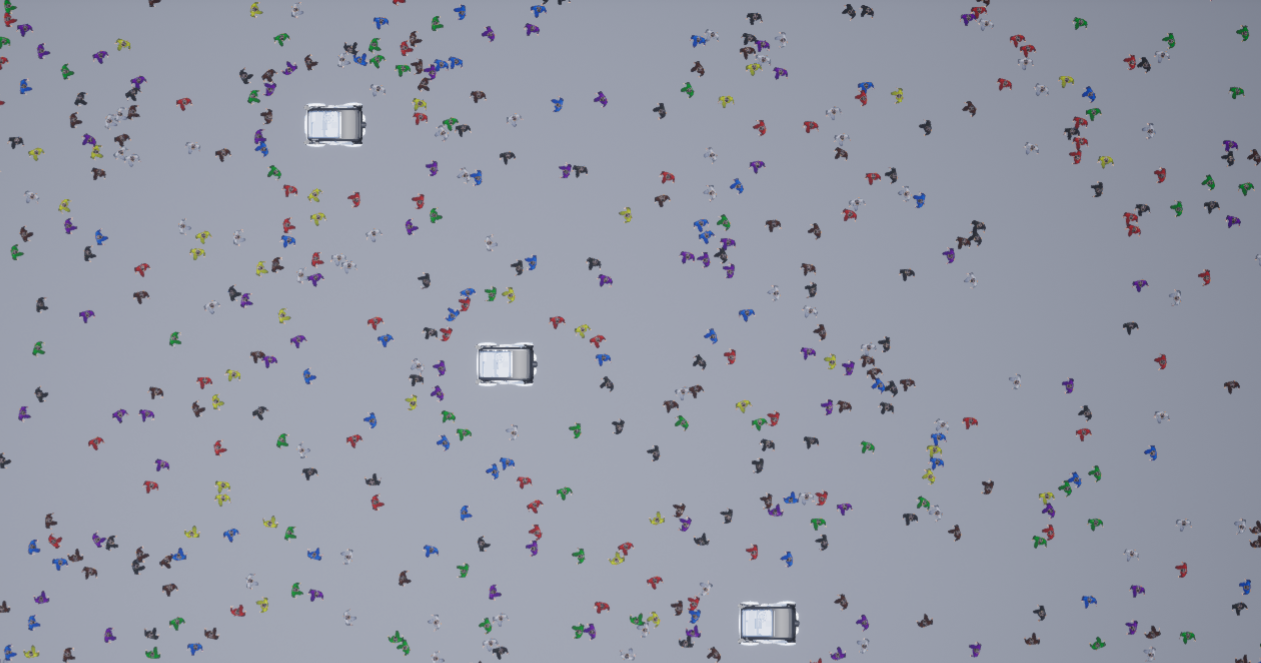}
        \caption{ }
    \end{subfigure}
     \begin{subfigure}{1\textwidth}
        \includegraphics[width=0.9\linewidth]{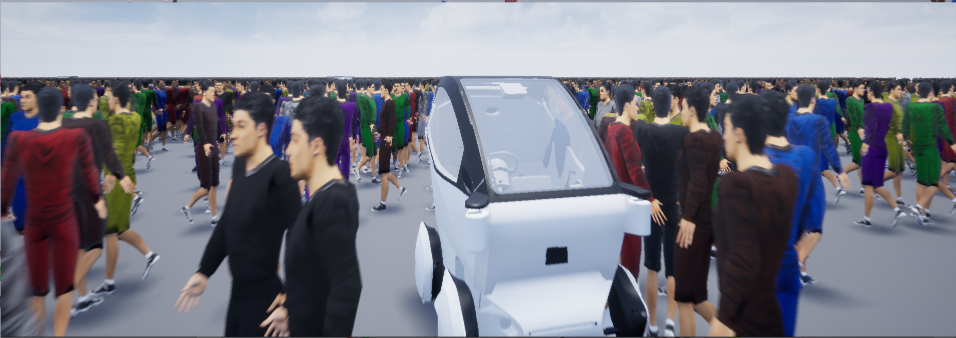}
        \caption{}
    \end{subfigure}
    \caption{Visualization of 2,500 agents in Unreal. Four crowds navigate past each other, two heading between left and right and two between top and bottom. a) Shows a zoomed-out view, with arrows showing the motion of each agent. b) shows a zoomed-in section of the above image. c) shows an eye-level view of the simulation }
    \label{fig:vis3}
\end{figure}

The first test case was a 2-way crossing, with the two crowds attempting to pass amongst each other to reach their destination. Figure \ref{fig:vis1} shows the result of this. 
The second test case was a 4-way crossing, visualized in figure \ref{fig:vis3}. Each crowd must navigate directly across the environment, causing avoidance of not only head-on agents, but also side-on. The 3D model of the vehicle is that of the ``Pod", used by the Transport Systems Catapult \cite{transport_systems_catapult_driverless_nodate}

\begin{figure}
    \centering
    \includegraphics[width=.88\linewidth]{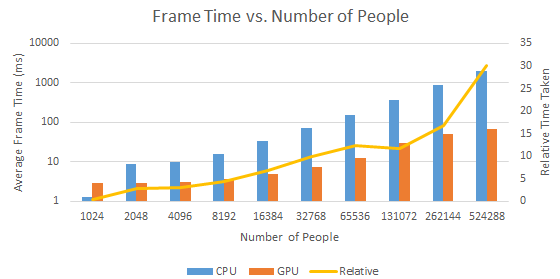}
    \caption{Frame time (in ms) for multi-core CPU and GPU ORCA models with varying numbers of agents. Logarithmic scale on primary (left) vertical axis. Relative timing is given on the secondary (right) vertical axis, in linear scale. Simulation time only without visualization of the pedestrians.}
    \label{fig:frameTime}
\end{figure}

The second experiment was designed to test the performance of the GPU implementation in comparison to the multi-core CPU implementation. Figure \ref{fig:frameTime} shows the results that varying numbers of agents have on the frame time. Various test cases (e.g. 2-way and 4-way crossings) with different agent parameters were run, and the timings averaged between them. In this experiment no visualization was used so as to ensure the timings were due to the algorithm only. The GPU solution gives speed increases of up to 30 times compared to the multi-core CPU implementation. Results for the single-core CPU version are not given as for any sizeable number of agents the multi-core CPU implementation always outperforms the single-core CPU implementation. This is due to better utilization of the CPU device. The colored bars of figure \ref{fig:frameTime} correspond to the primary (left) vertical axis, which uses a logarithmic scale. The relative time taken between the charts corresponds to the secondary (right) vertical axis, with linear scale. This result is the same that was found when simulating only people, as in the previous work of Charlton et al., using the same hardware \cite{charlton_fast_2019}. This occurs because there is no computational difference between agents representing people or vehicles. On the GPU this results in the same code being executed, which maintains parallelism across the device and results in consistent performance regardless of the proportions of people and autonomous cars. Hence, the same performance between the proposed model and the previous work is maintained.

The results show that the speed increases proportionally to the number of agents. Greater relative speed-up occurs for even larger numbers of agents, but the time taken per frame is below real time. The GPU simulations ran at close to 30 frames a second (33ms per frame) for up to $5\times10^5$ agents. The CPU version performs better for smaller number of agents, with a crossover occurring at approximately  $2\times10^3$ agents. This is due to the GPU device not being fully utilized for smaller simulations and the reduced throughput being outperformed by the CPU. The experiments were run on an NVIDIA GTX 970 GPU card with 4GB dedicated memory and a 4-core/8-thread Intel i7-4790K with 16 GB RAM. The GPU was connected by PCI-E 2.0. The GPU software was developed with NVIDIA CUDA 8.0 on Windows 10. On the GPU tested, there was a limit on the amount of usable memory of 4GB, which corresponded to approximately $5\times10^5$ agents. It is expected that relative performance increases will continue to be obtained for larger numbers of agents for the GPU implementation for GPUs with larger memory capacity. For example a modern Tesla V100 with 32GB of on-board RAM would support up to $4\times10^6$ agents.

\section{Conclusions} \label{sec:conclusion}

We have introduced extensions to the previously presented GPU ORCA model  \cite{charlton_fast_2019} to allow for simulation of autonomous vehicles. The flexibility in being able to alter the amount of avoidance people and vehicles make with each other allows for simulating a wide variety of different cases. The experimental results shows were for agents taking full responsibility for avoiding colliding with vehicles, however the model contains the ability to alter this amount. This aids autonomous vehicle designers to assess the levels of responsibility to achieve safe operation within a crowded environment.

Our model is currently limited in the number of agents in the simulation size due to GPU memory. The models use large amounts of memory for storing the ORCA half-planes of each person. Memory usage could be reduced by considering fewer people. This would reduce the memory of each person but may result in less realistic motion with greater chance of collisions.  A solution to the lack of memory is with Maxwell and later architectures, which can use managed memory \cite{nvidia_tuning_2018} to page information from CPU to GPU on demand.  This would allow for many more people to be simulated, up to the computer's system RAM capacity. It is expected that greater relative speedups between multi-core CPU and GPU will continue to be obtained for even larger amounts of simulated people.

It is expected that the more computationally expensive steering models would include more realistic motion such as side-stepping for pedestrians, more realistic densities, and less probability of collisions. In comparison, it is expected that the model in this paper would have greater performance and larger numbers of simulated people. An interesting extension would be to account for non-holonomic motion, such as the work of Alonso-Mora et al. \cite{alonso-mora_reciprocal_2012} or acceleration, Berg et al. \cite{berg_reciprocal_2011}. These models create extra linear constraints that arise due to the limitation in motion of the agent, and further constrain the possible motions. The simulation would then be composed of the ORCA model for pedestrians, and one of these extended models for vehicles, interacting and avoiding each other.

The current work involves writing the data from the simulation to a file before visualization using Unreal. The data is copied from the GPU to the CPU, then loaded into Unreal and copied back to the GPU in Unreal for visualization. This is expensive. Future work will look at how to use the Unreal engine to visualize a simulation as it is calculated, which could be done by sharing GPU buffer information between the simulation program and the Unreal Engine. 

A further extension to this model includes the use of virtual reality hardware. Users are placed within the virtual simulation to navigate around simulated people and vehicles. This could function as training for people to become accustomed to being in the same area as autonomous vehicles. The use of a treadmill or other walking device that combines with virtual reality would also provide benefits. By physically walking, it would create stronger immersion in the simulation.

\begin{acknowledgements}
This research was supported by EPSRC grant ``Accelerating Scientific Discovery with Accelerated Computing" (grant number EP/N018869/1),
and by the Transport Systems Catapult, and the National Council of Science and Technology in Mexico (Consejo Nacional de Ciencia y Tecnolog\'ia, CONACYT).
\end{acknowledgements}

%
%
%
\bibliographystyle{ieeetr}
\bibliography{MyLibraryBBT}

\begin{thebibliography}{10}

\bibitem{barut_combining_2018}
O.~Barut, M.~Haciomeroglu, and E.~A. Sezer, ``Combining {GPU}-generated linear
  trajectory segments to create collision-free paths for real-time ambient
  crowds,'' {\em Graphical Models}, vol.~99, pp.~31--45, Sept. 2018.

\bibitem{bleiweiss_multi_2009}
A.~Bleiweiss, ``Multi agent navigation on the gpu,'' {\em White paper, GDC},
  vol.~9, 2009.

\bibitem{charlton_fast_2019}
J.~Charlton, L.~R.~M. Gonzalez, S.~Maddock, and P.~Richmond, ``Fast
  {Simulation} of {Crowd} {Collision} {Avoidance},'' in {\em Computer
  {Graphics} {International} {Conference}}, pp.~266--277, Springer, 2019.

\bibitem{charlton_two-dimensional_2019}
J.~Charlton, S.~Maddock, and P.~Richmond, ``Two-dimensional batch linear
  programming on the {GPU},'' {\em Journal of Parallel and Distributed
  Computing}, vol.~126, pp.~152--160, Apr. 2019.

\bibitem{richmond_flame_2011}
P.~Richmond, {\em Flame gpu technical report and user guide}.
\newblock Department of {Computer} {Science} {Technical} {Report} {CS}-11-03,
  University of Sheffield, 2011.

\bibitem{pettre_motion_2008}
J.~Pettré, M.~Kallmann, and M.~C. Lin, ``Motion {Planning} and {Autonomy} for
  {Virtual} {Humans},'' in {\em {ACM} {SIGGRAPH} 2008 {Classes}}, {SIGGRAPH}
  '08, (New York, NY, USA), pp.~42:1--42:31, ACM, 2008.

\bibitem{bousseau_introduction_2017}
J.~Pettré and N.~Pelechano, ``Introduction to {Crowd} {Simulation},'' in {\em
  {EG} 2017 - {Tutorials}} (A.~Bousseau and D.~Gutierrez, eds.), The
  Eurographics Association, 2017.

\bibitem{thalmann_populating_2006}
D.~Thalmann, ``Populating {Virtual} {Environments} with {Crowds},'' in {\em
  Proceedings of the 2006 {ACM} {International} {Conference} on {Virtual}
  {Reality} {Continuum} and {Its} {Applications}}, {VRCIA} '06, (New York, NY,
  USA), pp.~11--11, ACM, 2006.
\newblock event-place: Hong Kong, China.

\bibitem{narain_aggregate_2009}
R.~Narain, A.~Golas, S.~Curtis, and M.~C. Lin, ``Aggregate {Dynamics} for
  {Dense} {Crowd} {Simulation},'' in {\em {ACM} {SIGGRAPH} {Asia} 2009
  {Papers}}, {SIGGRAPH} {Asia} '09, (New York, NY, USA), pp.~122:1--122:8, ACM,
  2009.

\bibitem{fickett_GPU_2007}
M.~FICKETT and L.~ZARKO, ``{GPU} {Continuum} {Crowds},'' {\em CIS Final Project
  Final report, University of Pennsylvania}, 2007.

\bibitem{helbing_social_1995}
D.~Helbing and P.~Molnár, ``Social force model for pedestrian dynamics,'' {\em
  Physical Review E}, vol.~51, pp.~4282--4286, May 1995.

\bibitem{fiorini_motion_1998}
P.~Fiorini and Z.~Shiller, ``Motion {Planning} in {Dynamic} {Environments}
  {Using} {Velocity} {Obstacles},'' {\em The International Journal of Robotics
  Research}, vol.~17, pp.~760--772, July 1998.

\bibitem{blue_emergent_1998}
V.~Blue and J.~Adler, ``Emergent {Fundamental} {Pedestrian} {Flows} {From}
  {Cellular} {Automata} {Microsimulation} {\textbar} {Request} {PDF},'' {\em
  Transportation Research Record: Journal of the Transportation Research
  Board}, vol.~1644, pp.~29--36, 1998.

\bibitem{blue_cellular_1999}
V.~Blue and J.~Adler, ``Cellular {Automata} {Microsimulation} of
  {Bidirectional} {Pedestrian} {Flows},'' {\em Transportation Research Record:
  Journal of the Transportation Research Board}, vol.~1678, pp.~135--141, Jan.
  1999.

\bibitem{schonfisch_synchronous_1999}
B.~Schönfisch and A.~de~Roos, ``Synchronous and asynchronous updating in
  cellular automata,'' {\em Biosystems}, vol.~51, pp.~123--143, Sept. 1999.

\bibitem{karmakharm_agent-based_2010}
T.~Karmakharm and P.~Richmond, ``Agent-based {Large} {Scale} {Simulation} of
  {Pedestrians} {With} {Adaptive} {Realistic} {Navigation} {Vector} {Fields},''
  {\em EG UK Theory and Practice of Computer Graphics}, 2010.

\bibitem{richmond_high_2008}
P.~Richmond and D.~M. Romano, ``A high performance framework for agent based
  pedestrian dynamics on gpu hardware,'' {\em Proceedings of EUROSIS ESM},
  vol.~2008, 2008.

\bibitem{abe_collision_2001}
Y.~Abe and M.~Yoshiki, ``Collision avoidance method for multiple autonomous
  mobile agents by implicit cooperation,'' in {\em Proceedings 2001
  {IEEE}/{RSJ} {International} {Conference} on {Intelligent} {Robots} and
  {Systems}. {Expanding} the {Societal} {Role} of {Robotics} in the the {Next}
  {Millennium} ({Cat}. {No}.{01CH37180})}, vol.~3, pp.~1207--1212 vol.3, Oct.
  2001.

\bibitem{kluge_recursive_2006}
B.~Kluge and E.~Prassler, ``Recursive {Probabilistic} {Velocity} {Obstacles}
  for {Reflective} {Navigation},'' in {\em Field and {Service} {Robotics}:
  {Recent} {Advances} in {Reserch} and {Applications}} (S.~Yuta, H.~Asama,
  E.~Prassler, T.~Tsubouchi, and S.~Thrun, eds.), Springer {Tracts} in
  {Advanced} {Robotics}, pp.~71--79, Berlin, Heidelberg: Springer Berlin
  Heidelberg, 2006.

\bibitem{fulgenzi_dynamic_2007}
C.~Fulgenzi, A.~Spalanzani, and C.~Laugier, ``Dynamic {Obstacle} {Avoidance} in
  uncertain environment combining {PVOs} and {Occupancy} {Grid},'' in {\em
  Proceedings 2007 {IEEE} {International} {Conference} on {Robotics} and
  {Automation}}, (Rome, Italy), pp.~1610--1616, IEEE, Apr. 2007.

\bibitem{berg_reciprocal_2008}
J.~v.~d. Berg, M.~Lin, and D.~Manocha, ``Reciprocal {Velocity} {Obstacles} for
  real-time multi-agent navigation,'' in {\em {IEEE} {International}
  {Conference} on {Robotics} and {Automation}, 2008. {ICRA} 2008},
  pp.~1928--1935, May 2008.

\bibitem{guy_clearpath:_2009}
S.~J. Guy, J.~Chhugani, C.~Kim, N.~Satish, M.~Lin, D.~Manocha, and P.~Dubey,
  ``{ClearPath}: {Highly} {Parallel} {Collision} {Avoidance} for {Multi}-agent
  {Simulation},'' in {\em Proceedings of the 2009 {ACM}
  {SIGGRAPH}/{Eurographics} {Symposium} on {Computer} {Animation}}, {SCA} '09,
  (New York, NY, USA), pp.~177--187, ACM, 2009.

\bibitem{he_dynamic_2016}
L.~He, J.~Pan, S.~Narang, W.~Wang, and D.~Manocha, ``Dynamic {Group}
  {Behaviors} for {Interactive} {Crowd} {Simulation},'' {\em arXiv:1602.03623
  [cs]}, Feb. 2016.
\newblock arXiv: 1602.03623.

\bibitem{yang_proxemic_2016}
Z.~Yang, J.~Pan, W.~Wang, and D.~Manocha, ``Proxemic group behaviors using
  reciprocal multi-agent navigation,'' {\em 2016 IEEE International Conference
  on Robotics and Automation (ICRA)}, pp.~292--297, 2016.

\bibitem{best_real-time_2016}
A.~Best, S.~Narang, and D.~Manocha, ``Real-time reciprocal collision avoidance
  with elliptical agents,'' in {\em 2016 {IEEE} {International} {Conference} on
  {Robotics} and {Automation} ({ICRA})}, pp.~298--305, IEEE, 2016.

\bibitem{hughes_holonomic_2014}
R.~Hughes, J.~Ondřej, and J.~Dingliana, ``Holonomic {Collision} {Avoidance}
  for {Virtual} {Crowds},'' in {\em Proceedings of the {ACM}
  {SIGGRAPH}/{Eurographics} {Symposium} on {Computer} {Animation}}, {SCA} '14,
  (Aire-la-Ville, Switzerland, Switzerland), pp.~103--111, Eurographics
  Association, 2014.

\bibitem{wang2018reciprocal}
L.~Wang, Z.~Li, C.~Wen, R.~He, and F.~Guo, ``Reciprocal collision avoidance for
  nonholonomic mobile robots,'' in {\em 2018 15th international conference on
  control, automation, robotics and vision ({ICARCV})}, pp.~371--376, 2018.
\newblock tex.organization: IEEE.

\bibitem{huang2019generalized}
X.~Huang, L.~Zhou, Z.~Guan, Z.~Li, C.~Wen, and R.~He, ``Generalized reciprocal
  collision avoidance for non-holonomic robots,'' in {\em 2019 14th {IEEE}
  conference on industrial electronics and applications ({ICIEA})},
  pp.~1623--1628, 2019.
\newblock tex.organization: IEEE.

\bibitem{snape_optimal_2019}
J.~Snape, ``Optimal {Reciprocal} {Collision} {Avoidance} ({C}++).
  github.com/snape/{RVO2},'' Mar. 2019.
\newblock original-date: 2013-06-25T03:11:32Z.

\bibitem{douthwaite2019velocity}
J.~A. Douthwaite, S.~Zhao, and L.~S. Mihaylova, ``Velocity obstacle approaches
  for multi-agent collision avoidance,'' {\em Unmanned Systems}, vol.~7,
  no.~01, pp.~55--64, 2019.
\newblock tex.publisher: World Scientific Publishing Company.

\bibitem{seidel_small-dimensional_1991}
R.~Seidel, ``Small-dimensional linear programming and convex hulls made easy,''
  {\em Discrete \& Computational Geometry}, vol.~6, pp.~423--434, Sept. 1991.

\bibitem{gogoll_autonomous_2017}
J.~Gogoll and J.~F. Müller, ``Autonomous cars: in favor of a mandatory ethics
  setting,'' {\em Science and engineering ethics}, vol.~23, no.~3,
  pp.~681--700, 2017.

\bibitem{lin_why_2016}
P.~Lin, ``Why ethics matters for autonomous cars,'' in {\em Autonomous
  driving}, pp.~69--85, Springer, Berlin, Heidelberg, 2016.

\bibitem{mcbride_ethics_2016}
N.~McBride, ``The {Ethics} of {Driverless} {Cars},'' {\em SIGCAS Comput. Soc.},
  vol.~45, pp.~179--184, Jan. 2016.

\bibitem{althoff_comparison_2011}
M.~Althoff and A.~Mergel, ``Comparison of {Markov} chain abstraction and
  {Monte} {Carlo} simulation for the safety assessment of autonomous cars,''
  {\em IEEE Transactions on Intelligent Transportation Systems}, vol.~12,
  no.~4, pp.~1237--1247, 2011.

\bibitem{guo_is_2019}
J.~Guo, U.~Kurup, and M.~Shah, ``Is {It} {Safe} to {Drive}? {An} {Overview} of
  {Factors}, {Metrics}, and {Datasets} for {Driveability} {Assessment} in
  {Autonomous} {Driving},'' {\em IEEE Transactions on Intelligent
  Transportation Systems}, 2019.

\bibitem{rasouli_autonomous_2019}
A.~Rasouli and J.~K. Tsotsos, ``Autonomous vehicles that interact with
  pedestrians: {A} survey of theory and practice,'' {\em IEEE transactions on
  intelligent transportation systems}, 2019.

\bibitem{schwarting_social_2019}
W.~Schwarting, A.~Pierson, J.~Alonso-Mora, S.~Karaman, and D.~Rus, ``Social
  behavior for autonomous vehicles,'' {\em Proceedings of the National Academy
  of Sciences}, vol.~116, no.~50, pp.~24972--24978, 2019.

\bibitem{bonnefon_social_2016}
J.-F. Bonnefon, A.~Shariff, and I.~Rahwan, ``The social dilemma of autonomous
  vehicles,'' {\em Science}, vol.~352, no.~6293, pp.~1573--1576, 2016.

\bibitem{pettersson_setting_2015}
I.~Pettersson and I.~C.~M. Karlsson, ``Setting the stage for autonomous cars: a
  pilot study of future autonomous driving experiences,'' {\em IET Intelligent
  Transport Systems}, vol.~9, pp.~694--701, May 2015.

\bibitem{boesch_agent-based_2015}
P.~M. Boesch and F.~Ciari, ``Agent-based simulation of autonomous cars,'' in
  {\em 2015 {American} {Control} {Conference} ({ACC})}, pp.~2588--2592, IEEE,
  2015.

\bibitem{horl_agent-based_2017}
S.~Hörl, ``Agent-based simulation of autonomous taxi services with dynamic
  demand responses,'' {\em Procedia Computer Science}, vol.~109, pp.~899--904,
  2017.

\bibitem{zhang_exploring_2015}
W.~Zhang, S.~Guhathakurta, J.~Fang, and G.~Zhang, ``Exploring the impact of
  shared autonomous vehicles on urban parking demand: {An} agent-based
  simulation approach,'' {\em Sustainable Cities and Society}, vol.~19,
  pp.~34--45, 2015.

\bibitem{berg_reciprocal_2011}
J.~v.~d. Berg, J.~Snape, S.~J. Guy, and D.~Manocha, ``Reciprocal collision
  avoidance with acceleration-velocity obstacles,'' in {\em 2011 {IEEE}
  {International} {Conference} on {Robotics} and {Automation}}, pp.~3475--3482,
  May 2011.

\bibitem{wang_gunrock:_2016}
Y.~Wang, A.~Davidson, Y.~Pan, Y.~Wu, A.~Riffel, and J.~D. Owens, ``Gunrock: {A}
  {High}-{Performance} {Graph} {Processing} {Library} on the {GPU},'' {\em
  arXiv:1501.05387 [cs]}, pp.~1--12, 2016.
\newblock arXiv: 1501.05387.

\bibitem{transport_systems_catapult_driverless_nodate}
T.~S. Catapult, ``Driverless {Pods}.''

\bibitem{nvidia_tuning_2018}
Nvidia, ``Tuning {CUDA} {Applications} for {Maxwell},'' 2018.

\bibitem{alonso-mora_reciprocal_2012}
J.~Alonso-Mora, A.~Breitenmoser, P.~Beardsley, and R.~Siegwart, ``Reciprocal
  collision avoidance for multiple car-like robots,'' in {\em 2012 {IEEE}
  {International} {Conference} on {Robotics} and {Automation}}, pp.~360--366,
  May 2012.

\end{thebibliography}

\end{document}